\documentclass[letterpaper]{article} 
\usepackage{aaai2026}  
\usepackage{times}  
\usepackage{helvet}  
\usepackage{courier}  
\usepackage[hyphens]{url}  
\usepackage{graphicx} 
\usepackage{natbib}  
\usepackage{caption} 
\usepackage{booktabs}
\usepackage{multirow}
\usepackage{amssymb}
\usepackage{pifont}
\usepackage{enumitem}
\usepackage{tabularx}

\usepackage{algorithm}
\usepackage{algorithmic}
\usepackage[table,xcdraw]{xcolor}

\usepackage{newfloat}
\usepackage{listings}
\DeclareCaptionStyle{ruled}{labelfont=normalfont,labelsep=colon,strut=off} 
\floatstyle{ruled}
\newfloat{listing}{tb}{lst}{}
\floatname{listing}{Listing}
\title{AirCopBench: A Benchmark for Multi-drone Collaborative \\
Embodied Perception and Reasoning}
\author{
    Jirong Zha \textsuperscript{\rm 1}\textsuperscript{*$\blacklozenge$}, Yuxuan Fan\textsuperscript{\rm 2}\textsuperscript{*},
    Tianyu Zhang\textsuperscript{\rm 3}, Geng Chen\textsuperscript{\rm 4}, 
    \\
    Yingfeng Chen\textsuperscript{\rm 5}, Chen Gao\textsuperscript{\rm 6 \dag}, Xinlei Chen\textsuperscript{\rm 1 \dag}
}
\affiliations{
    \textsuperscript{\rm 1}Shenzhen International Graduate School, Tsinghua University \\
    \textsuperscript{\rm 2}The Hong Kong University of Science and Technology (Guang Zhou) \\
    \textsuperscript{\rm 3}School of Electrical and Electronic Engineering, Nanyang Technological University \\
    \textsuperscript{\rm 4}College of Software, Jilin University \\
    \textsuperscript{\rm 5}Weiyang College, Tsinghua University \\
    \textsuperscript{\rm 6}BNRist, Tsinghua University \\
    zhajirong23@mails.tsinghua.edu.cn, yfan546@connect.hkust-gz.edu.cn,
    tianyu016@e.ntu.edu.sg,
    chengeng0201@gmail.com,
    chenying24@mails.tsinghua.edu.cn,
    chgao96@gmail.com,
    chen.xinlei@sz.tsinghua.edu.cn \\
    \textsuperscript{*} Equal Contribution.
    \textsuperscript{$\blacklozenge$} Project Leader.
    \textsuperscript{\dag} Corresponding Authors.
}

\begin{document}

\makeatletter 

\twocolumn[{%
\begin{center}
    \vskip 0.625in minus 0.125in

    {\LARGE\bf \@title \par}%

    \vskip 0.1in plus 0.5fil minus 0.05in

    \def\theauthors{\if T\showauthors@on\@author\else Anonymous submission\fi}

    {\Large{\textbf{\theauthors}}\par}%

    \vskip .2em plus 0.25fil

    {\normalsize \affiliations_\par}%

    \vskip 1em plus 2fil

    \includegraphics[width=1\linewidth]{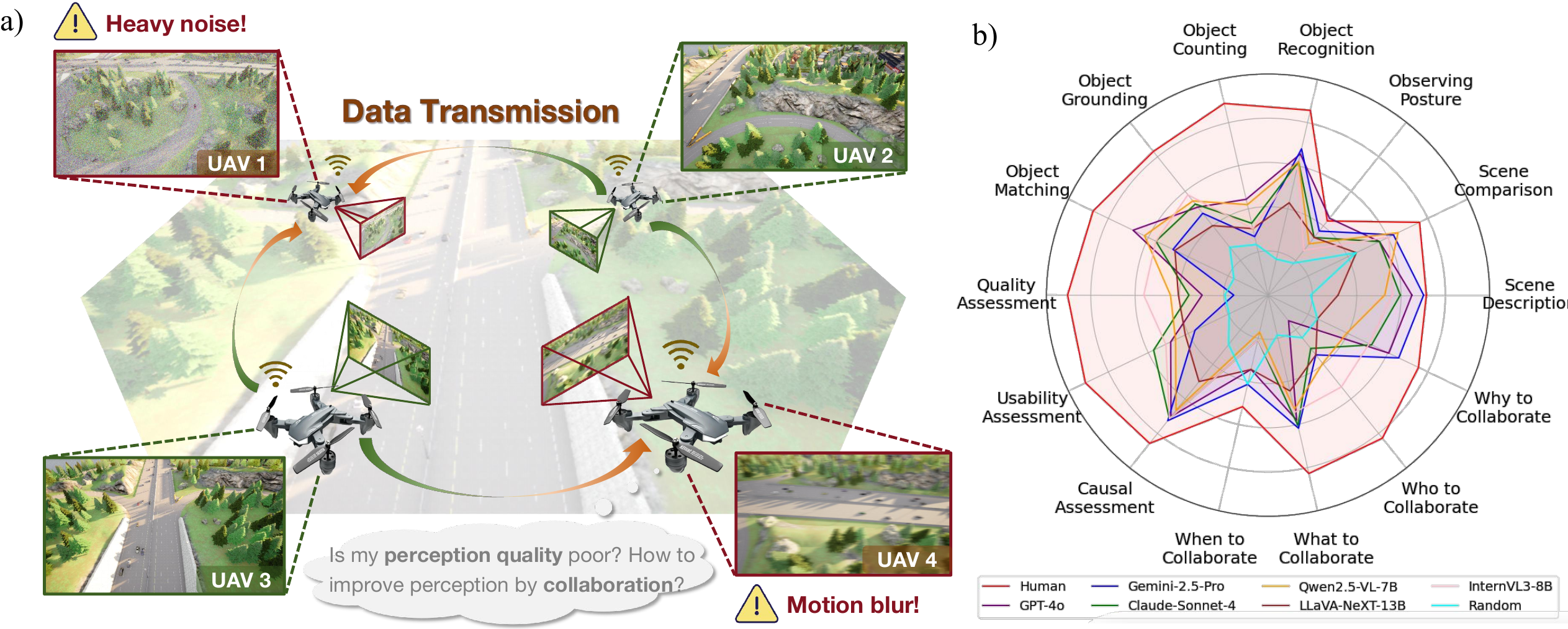}
    \captionof{figure}{a) Illustration of multi-drone collaborative perception with various perception degradation. b) The performance of 6 popular MLLMs, along with human and random guess baselines, on AirCopBench.}
    \label{intro}
\end{center}
}]

\makeatother


\begin{abstract}
Multimodal Large Language Models (MLLMs) have shown promise in single-agent vision tasks, yet benchmarks for evaluating multi-agent collaborative perception remain scarce. This gap is critical, as multi-drone systems provide enhanced coverage, robustness, and collaboration compared to single-sensor setups. Existing multi-image benchmarks mainly target basic perception tasks using high-quality single-agent images, thus failing to evaluate MLLMs in more complex, egocentric collaborative scenarios, especially under real-world degraded perception conditions.To address these challenges, we introduce AirCopBench, the first comprehensive benchmark designed to evaluate MLLMs in embodied aerial collaborative perception under challenging perceptual conditions. AirCopBench includes 14.6k+ questions derived from both simulator and real-world data, spanning four key task dimensions: Scene Understanding, Object Understanding, Perception Assessment, and Collaborative Decision, across 14 task types. 
We construct the benchmark using data from challenging degraded-perception scenarios with annotated collaborative events, generating large-scale questions through model-, rule-, and human-based methods under rigorous quality control. Evaluations on 40 MLLMs show significant performance gaps in collaborative perception tasks, with the best model trailing humans by 24.38\% on average and exhibiting inconsistent results across tasks. Fine-tuning experiments further confirm the feasibility of sim-to-real transfer in aerial collaborative perception and reasoning.
\end{abstract}

\begin{links}
\vspace{-0.3cm}
    \link{Project}{https://embodiedcity.github.io/AirCopBench/}
    \vspace{-0.3cm}
\end{links}

\begin{table*}[t]
\centering
\caption{Comparison of the proposed and popular benchmarks for collaborative perception.}
\resizebox{\textwidth}{!}
{
\begin{tabular}{lcccccccc}
\toprule
\textbf{Benchmark} & \textbf{Modality} & \textbf{Object Category} & \textbf{Perception Degradation} & \textbf{VQA} & \textbf{VQA Num.} & \textbf{Embodied} \\ 
\midrule
Coperception-UAV \cite{hu2022where2comm} & RGB & Vehicles & / & \ding{55} & / & \ding{51} \\
MDMT \cite{liu2023robust} & RGB & Vehicles, bicycles, pedestrians & Occlusion, lighting, blur & \ding{55} & / & \ding{51} \\
UAV3D \cite{sunderraman2024uav3d} & RGB & Vehicles & / & \ding{55} & / & \ding{51} \\
AeroCollab3D \cite{tian2024ucdnet} & RGB & Vehicles, pedestrians & / & \ding{55} & / & \ding{51} \\
Air-Co-Pred \cite{wang2024drones} & RGB & Vehicles, bicycles, pedestrians & Occlusion, long distance & \ding{55} & / & \ding{51} \\
MUIRBENCH \cite{wang2024muirbench} & RGB, text & Vehicles, buildings & Occlusion & \ding{51} & 2.6k & \ding{55}\\
All-Angles Bench \cite{yeh2025seeing} & RGB, text & Pedestrians & Occlusion & \ding{51} & 2.1k & \ding{55} \\
UrBench \cite{zhou2025urbench} & RGB, text & Vehicles, buildings & / & \ding{51} & 11.6k & \ding{55} \\
\midrule
\multirow{2}{*}{Our AirCopBench} & RGB, text, & Drones, vehicles, bicycles, & Occlusion, shadow, noise, lighting, out of FoV,  & \multirow{2}{*}{\ding{51}} & \multirow{2}{*}{14.6k} & \multirow{2}{*}{\ding{51}} \\
 & point cloud & pedestrians & data loss, long distance/small target, blur &  & \\
\bottomrule
\end{tabular}
}
\label{comparison}
\end{table*}

\section{Introduction}

Multimodal Large Language Models (MLLMs) have transformed artificial intelligence (AI) by enabling powerful processing of diverse inputs, such as text and images \cite{caffagni2024revolution, zhang2024mm, fu2024blink}. While MLLMs have excelled in single-agent vision tasks like object detection and semantic segmentation \cite{saini2025advancing, zang2025contextual, huang2025mllm}, as well as general multi-image understanding tasks \cite{cheng2025evaluating, bai2025hallucination}, their performance in collaborative perception with multiple unmanned aerial vehicles (UAVs) remains underexplored. In multi-UAV systems, drones work together to tackle complex, dynamic visual tasks through information exchange, offering enhanced coverage, robustness, and flexibility compared to single-agent configurations \cite{liu2020when2com, zha2024diffusion, lin2025mcop}.

Despite the potential of multi-UAV systems, benchmarks evaluating MLLMs in collaborative perception are scarce. Current evaluations focus primarily on single-agent vision tasks, failing to address the unique challenges of multi-UAV collaborative perception, such as obstacle occlusion \cite{kil2024mllm, xiao2025uav, guo2025bedi, chen2025large}.
Furthermore, existing benchmarks on collaborative perception \cite{wang2024drones, tian2024ucdnet, liu2020when2com} in Tab.~\ref{comparison} face two major limitations that hinder their applicability in adaptive real-world scenarios:

\begin{itemize}[left=0pt]
    \item \textit{Oversimplified Perception Setups}:
    Despite occlusion, numerous factors such as sensor noise, poor visibility, data loss, and environmental interference can degrade the quality of UAVs' observation data, thus reducing overall perception accuracy. Current benchmarks consider only a limited set of degradation types, restricting their adaptability to real-world conditions.

    
    \item  \textit{Lack of Embodied Reasoning}: 
    Aerial collaborative perception demands rich semantic information processing and highly adaptive cooperation in complex environments. Traditional non-egocentric, programmatic decision-making schemes, based on global data fusion, hinder UAVs from making context-aware, human-like, first-view decisions, limiting their ability to understand their state, adapt to changes, and collaborate efficiently.

\end{itemize}
Therefore, it is essential to assess MLLMs' cognitive abilities in embodied perception and collaborative decision-making under challenging perceptual conditions, as illustrated in Fig.~\ref{intro}. More related work can be found in \textit{Appendix}.

Accordingly, we propose \textbf{AirCopBench}, a novel benchmark designed to evaluate MLLMs in multi-UAV collaborative perception under various challenging perception conditions. Specifically, we introduce an innovative task set comprising 14 tasks across 4 dimensions from simultaneous multi-view images. Our dataset includes challenging data from both simulators \cite{gao2024embodiedcity, hu2022where2comm, tian2024ucdnet} and real world \cite{liu2023robust}, collected from diverse UAV groups in representative degraded scenarios.
Besides traditional object-level labeling for perception tasks, we incorporate event-level labeling by human annotators for specific cooperation events. Then, we develop a general pipeline to generate high-quality visual question answering (VQA) pairs using model-based, rule-based, and human-based approaches, followed by stringent quality control measures, including standard review, blind filtering, and human refinement.
Finally, we evaluate our dataset on popular MLLMs in zero-shot settings, including both proprietary and open-source models, and conduct supervised fine-tuning (SFT) on Qwen2.5-series \cite{bai2025qwen2} and LLaVA-NeXT series \cite{liu2024llavanext} to validate the dataset's effectiveness. The results show that current MLLMs struggle with multi-view collaborative perception, exhibiting inconsistent performance across different task types.
%
\begin{figure*}[t]
    \centering
    \includegraphics[width=0.92\linewidth]{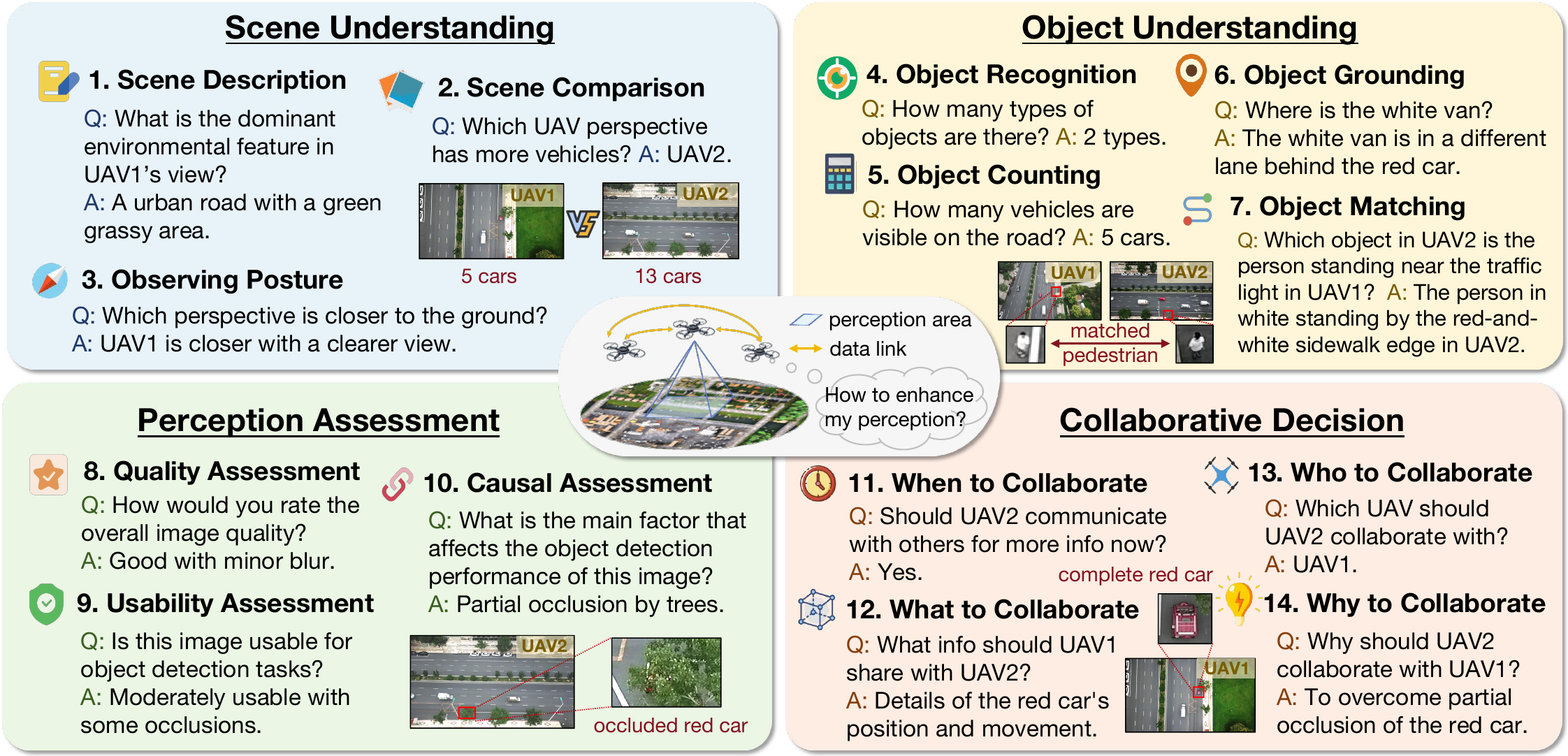}
    \caption{AirCopBench includes 14 task types across 4 evaluation dimensions: Scene Understanding, Object Understanding, Perception Assessment, and Collaborative Decision. This categorization facilitates a systematic evaluation of MLLMs, from image understanding and quality analysis to multi-UAV information exchange for improved collaborative embodied perception.}
    \label{task}
    \vspace{-0.3cm}
\end{figure*}

Compared to existing aerial collaborative perception benchmarks, key features of AirCopBench include: 
1) Semantic VQA pairs to evaluate MLLMs’ ability in perception assessment and collaborative reasoning;
2) Various challenging perception degradation in real-world scenarios, such as occlusion, shadows, motion blur, noise, data loss, long-range detection, complex background, etc;
3) Multiple modalities, including RGB images, text, and point clouds, to support perception across diverse data types;
4) Diverse target categories, covering drones, pedestrians, vehicles, and bicycles;
5) Embodied multi-UAV collaboration from first-view, role-based reasoning for human-like decision-making. 


Overall, the novelty of this research lies in creating \textbf{the first benchmark for semantic embodied aerial collaborative perception considering various challenging perception degradation}. Our contributions are fourfold:
\begin{itemize}[left=0pt]
    \item We introduce a novel task set with 4 categories and 14 tasks to evaluate MLLMs' abilities in scene and object understanding, perception assessment, and collaborative decision-making using multi-view images.
    \item We create an aerial collaborative perception dataset with over 2.9k+ multi-view images from UAV groups of varying sizes, annotated for collaborative events, and focused on diverse perception degradations.
    \item We construct 14.6k+ VQA pairs from collaborative UAV observations using both real and simulated data, and design an extensible benchmark pipeline applicable to other multi-view embodied settings.
    \item We evaluate 40 MLLMs and investigate the relationship between embodied collaborative perception tasks and Sim-to-Real potential. We also fine-tune 4 models to demonstrate the effectiveness of our dataset.
\end{itemize}

\section{AirCopBench}

AirCopBench is a high-quality aerial perception benchmark with 14 tasks across 4 dimensions, including challenging perception degradation scenarios. Based on quantitative multi-view images from various drone groups, it evaluates MLLMs' embodied collaborative perception capabilities. This section outlines the task definitions, generation pipeline, and statistical properties of the benchmark.



\subsection{Benchmark Tasks}

AirCopBench includes four key task dimensions: Scene Understanding, Object Understanding, Perception Assessment, and Collaborative Decision. Each category is further divided into sub-tasks to enable a more detailed evaluation of MLLMs' capabilities, as shown in Fig.~\ref{task}.


\subsubsection{Scene Understanding} refers to interpreting and understanding scenes from multi-view images (Task 1-3 in Fig.~\ref{task}). Specifically, Scene Description identifies objects, relationships, and context in one scene, while Scene Comparison reveals similarities and differences between images from different views \cite{zhou2025urbench}.
Observing Posture analyzes the relative distances and directions between observers and the scene, crucial for simulating human-like observations in embodied aerial perception tasks.

\subsubsection{Object Understanding} focuses on analyzing objects in terms of identity, quantity, and spatial relationships (Tasks 4-7 in Fig.~\ref{task}). Particularly, Object Recognition identifies object types, Object Counting determines object quantity for specific types, Object Grounding assesses the spatial positioning between objects, and Object Matching finds corresponding objects across different views \cite{zhou2025urbench}.

\subsubsection{Perception Assessment} evaluates image quality and usability for specific perception tasks (Task 8-10 in Fig.~\ref{task}). In detail, Quality Assessment rates image by objectively checking its clarity and resolution, while Usability Assessment subjectively determines its effectiveness for object detection. Causal Assessment traces the reasons for poor-quality or unsuitable images, analyzing whether degradation is due to sensor issues, moving targets, or environmental factors. This helps UAVs better understand their perception state and supports information exchange with other observing UAVs.

\begin{figure*}[t]
    \centering
    \includegraphics[width=0.95\linewidth]{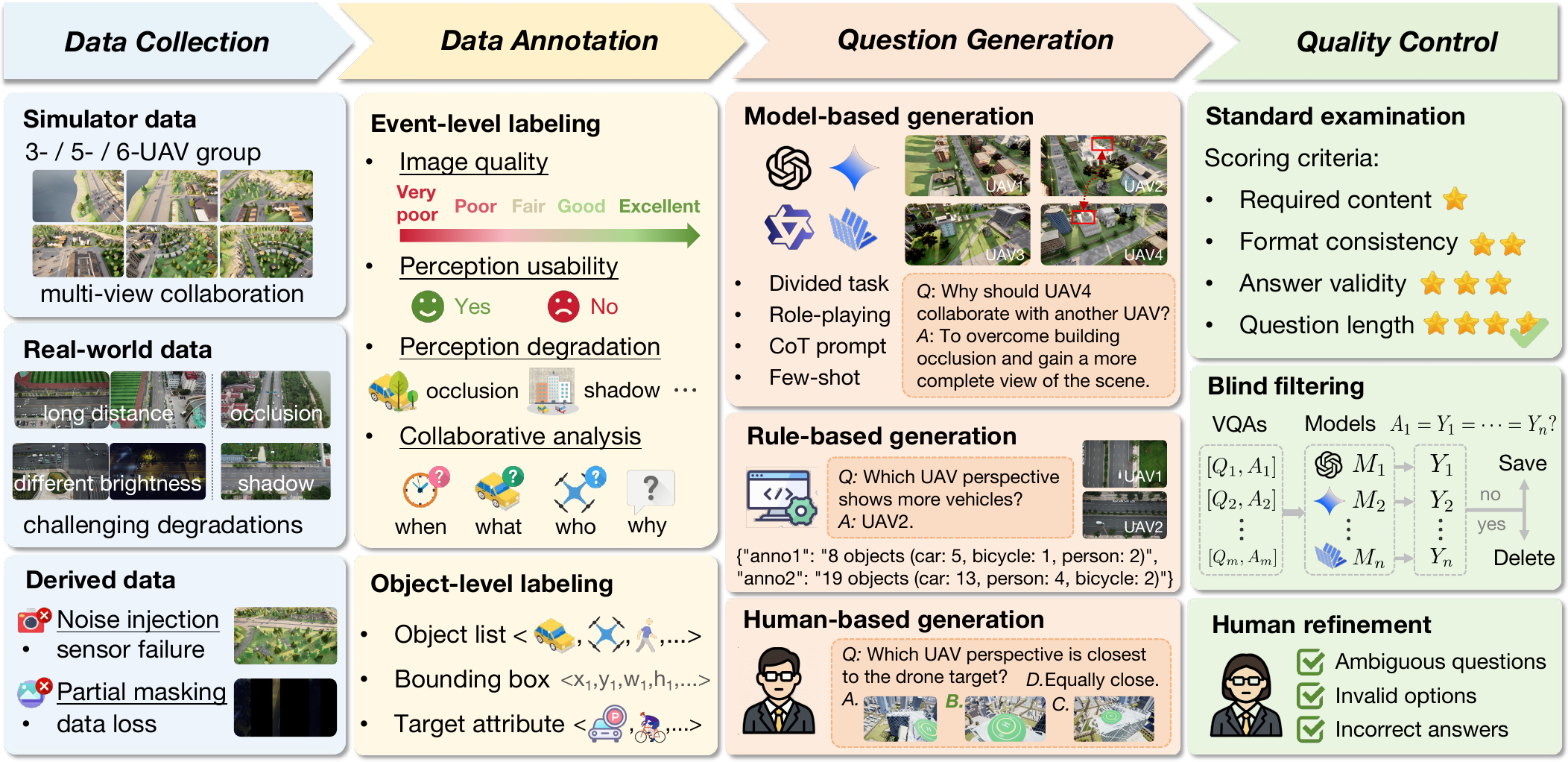}
    \caption{AirCopBench generation pipeline includes 4 main steps: Data Collection, Data Annotation, Question Generation, and Quality Control. This systematic approach ensures the validity and high quality of the generated dataset.}
    \label{pipeline}
    \vspace{-0.3cm}
\end{figure*}

\subsubsection{Collaborative Decision} determines the need for information exchange among UAVs (Task 11-14 in Fig.~\ref{task}). 
This task design aims to enhance collaboration efficiency between UAVs by reducing communication and computational costs of the multi-UAV system. 
Specifically, When to Collaborate identifies scenarios where collaboration is essential for current UAV, while What to Collaborate determines the information that should be shared between UAVs. Who to Collaborate assesses which UAVs are best suited for collaboration, and Why to Collaborate explores the motivations behind information exchange between UAVs.

\subsection{Benchmark Generation}
The benchmark generation pipeline consists of four main steps: Data Collection, Data Annotation, Question Generation, and Quality Control, as illustrated in Fig.~\ref{pipeline}.

\subsubsection{Data Collection.}
To effectively evaluate MLLMs in challenging scenarios, high-quality data with realistic perception degradations, such as occlusion, shadow, noise, lighting imbalance, data loss, motion blur, long distance, and out-of-field of view (FoV), is essential.
Our data consists of:
\\
\indent \textbf{Simulator Data.}
    Multimodal collaborative perception data, including RGB images and point clouds, is collected from co-simulated Carla \cite{dosovitskiy2017carla} and AirSim \cite{shah2017airsim}. This includes challenging ground vehicle target perception from 5-6 UAVs in existing datasets, Coperception-UAV \cite{hu2022where2comm} and AeroCollab3D \cite{tian2024ucdnet}, and drone target observation from 3 UAVs collected from EmbodiedCity \cite{gao2024embodiedcity}. The data spans 20 different map scenarios, focusing on those inducing key perceptual degradation issues, as detailed in \textit{Appendix}.
\\
\indent 
\textbf{Real-world Data.}
    To enhance data credibility, we select representative images from existing real-world dataset MDMT \cite{liu2023robust}, where 2 UAVs cooperatively perceive ground targets like vehicles, pedestrians, and bicycles. These scenarios include various perception challenges such as motion blur, varying lighting (day/night), distant small targets, and obstacle occlusions. 
\\
\indent
\textbf{Derived Data.}
    To expand the dataset and introduce more perceptual degradation, we apply two image post-processing techniques:
    a) Noise Injection adds random noises (Gaussian, salt-and-pepper, impulse, Poisson) to simulate sensor failures;
    b) Partial Masking blocks part of the image to simulate visual information loss.
    These techniques simulate perception degradation scenarios, such as low signal-to-noise ratio (SNR) and data loss due to sensor damage or failures, that are hard to capture in simulators or real environments.

\subsubsection{Data Annotation.}
Rich, rational annotations tailored for aerial collaborative perception improve dataset accuracy and realism. In this work, we focus on both event-level labels, capturing inter-UAV collaboration, and object-level labels, ensuring precise identification in complex scenes.
\\ \indent
\textbf{Event-level Labeling.} To emphasize the role of collaboration in perception, we introduce novel ``event" annotations to assist in generating multi-UAV 
interaction strategies,  
such as when and why inter-UAV communication is needed, who is suitable for information retrieval, and what observation information should be shared. Despite labeling for collaborative decision analysis, the annotations in this part also include image quality scoring, perception usability assessment, and perception degradation reasoning for better event understanding. The whole manual annotation process costs over 200 hours. More details are in \textit{Appendix}.
\\ \indent
\textbf{Object-level Labeling.} Traditional annotations for object labeling involve the list of specified objects in the scene, the 2D/3D bounding box of each object, and the corresponding attributes of objects like motion state \cite{hu2022where2comm}. This enables accurate collaborative perception for targets, including target detection, classification, and tracking.


 

\begin{figure*}[t]
    \centering
    \includegraphics[width=1\linewidth]{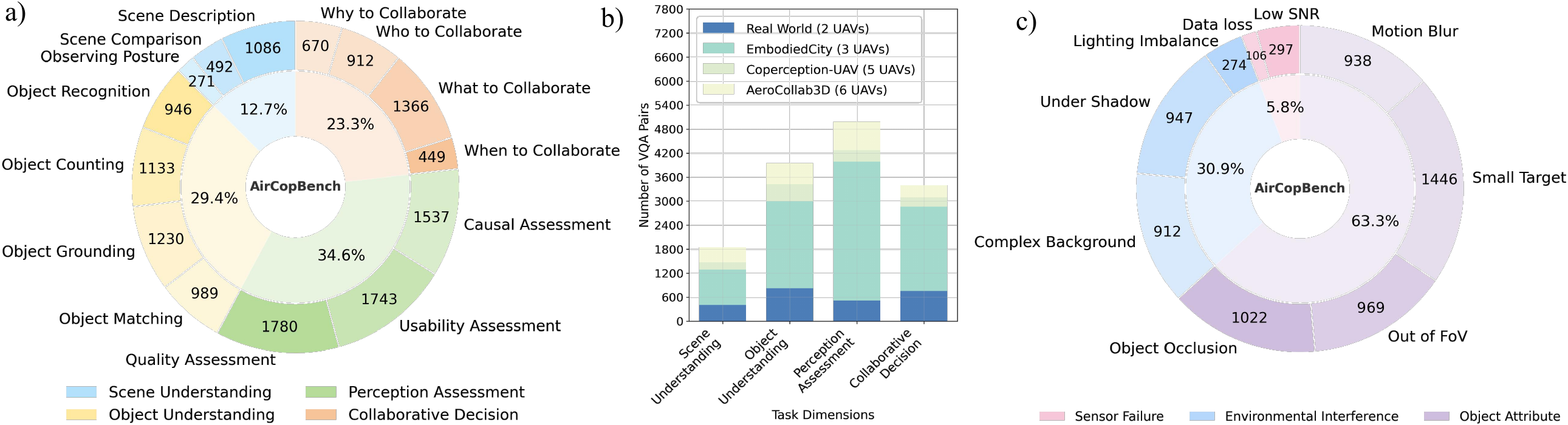}
    \caption{\textbf{Statistical Overview of AirCopBench.} a) Distribution of VQA pairs across 14 task types. b) Distribution of VQA pairs from various data sources with different numbers of observing UAV groups. c) Distribution of images featuring diverse perception degradation types.}
    \label{statistics}
    \vspace{-0.3cm}
\end{figure*}

\subsubsection{Question Generation.}
Given diverse complexity and requirements for different collaborative perception tasks, we design three approaches for question generation:
\\ \indent
\textbf{Model-based Generation.} 
This method uses powerful MLLMs, including GPT-4o \cite{hurst2024gpt} and Qwen-VL-Max-latest \cite{Qwen-VL}, to efficiently generate high-quality VQA pairs for each task. We employ four prompting techniques:
a) Task Decomposition breaks large tasks into smaller, focused tasks for better model understanding and easier debugging;
b) Role-playing Settings prompts models to generate questions from the perspective of different UAV agents in embodied perception tasks;
c) Chain-of-Thought (CoT) Prompting uses multi-step reasoning to generate complex questions requiring deep visual understanding;
d) Few-shot Learning leverages templates and examples for improved task understanding and generalization.
\\ \indent
\textbf{Rule-based Generation.}
This approach uses predefined rules and logic to generate structured questions. Tasks like Object Counting and Usability Assessment rely on rule-based methods, where questions are directly based on dataset annotations. These simple, deterministic rules make rule-based generation ideal for tasks requiring consistent, structured questions without complex reasoning or context.
\\ \indent
\textbf{Human-based Generation.}
This method supports tasks like Observing Posture and Object Matching, which require complex multi-image reasoning and spatial understanding. Expert annotations ensure subtle visual cues, realistic dynamics, and logical nuances are well captured, yielding high-quality, context-rich questions.




\begin{table*}[t]
\centering
\caption{Results on AirCopBench for existing various MLLMs on 14 task types across 4 evaluation dimensions. The best-performing model in each category is highlighted \textbf{in-bold}, while the second-best is \underline{underlined}. 24 out of 40 models and 3 out of 4 fine-tuned models for demonstration in the main text; additional results are provided in the \textit{Appendix}.}
\label{mllms_accuracy}
\renewcommand{\arraystretch}{1.2} 
\resizebox{\textwidth}{!}{
\begin{tabular}{%
  c|cc|
  *{3}{>{\centering\arraybackslash}p{0.8cm}}|
  *{4}{>{\centering\arraybackslash}p{0.8cm}}|
  *{3}{>{\centering\arraybackslash}p{0.8cm}}|
  *{4}{>{\centering\arraybackslash}p{0.8cm}}  
}

\hline
& & & 
\multicolumn{3}{c|}{\cellcolor{blue!10}\textbf{Scene Understanding}} 
& \multicolumn{4}{c|}{\cellcolor{yellow!10}\textbf{Object Undstanding}} 
& \multicolumn{3}{c|}{\cellcolor{green!10}\textbf{Perception Assessment}} 
& \multicolumn{4}{c}{\cellcolor{red!10}\textbf{Collaborative Decision}} \\
\multicolumn{1}{r|}{Method} 
& Rank 
& \multicolumn{1}{c|}{Avg.} 
& \rotatebox{90}{\textit{Scene Desc.}} 
& \rotatebox{90}{\textit{Scene Comp.}} 
& \rotatebox{90}{\textit{Obs. Post.}} 
& \rotatebox{90}{\textit{Obj. Rec.}} 
& \rotatebox{90}{\textit{Obj. Cnt.}} 
& \rotatebox{90}{\textit{Obj. Grnd.}} 
& \rotatebox{90}{\textit{Obj. Mtch.}} 
& \rotatebox{90}{\textit{Qual. Ass.}} 
& \rotatebox{90}{\textit{Usab. Ass.}} 
& \rotatebox{90}{\textit{Caus. Ass.}} 
& \rotatebox{90}{\textit{When Coll.}} 
& \rotatebox{90}{\textit{What Coll.}} 
& \rotatebox{90}{\textit{Who Coll.}} 
& \rotatebox{90}{\textit{Why Coll.}} \\
\hline
\rowcolor{cyan!8}
\multicolumn{1}{l|}{\textbf{\textit{Baseline}}} & \multicolumn{2}{l|}{} & \multicolumn{3}{l|}{} & \multicolumn{4}{l|}{} & \multicolumn{3}{l|}{} & \multicolumn{4}{l}{} \\
\multicolumn{1}{r|}{Random} & - & 23.47 & 19.30 & 44.19 & 18.52 & 16.67 & 23.46 & 27.68 & 17.14 & 19.51 & 19.51 & 28.57 & 41.38 & 18.52 & 24.69 & 24.69 \\
\multicolumn{1}{r|}{Human} & - & 78.25 & 71.43 & 75.86 & 42.86 & 85.71 & 88.89 & 83.04 & 87.62 & 90.48 & 91.46 & 85.71 & 51.72 & 82.72 & 82.72 & 75.31 \\
\hline
\rowcolor{cyan!8}
\multicolumn{1}{l|}{\textbf{\textit{Proprietary Models (API)}}} & \multicolumn{2}{l|}{} & \multicolumn{3}{l|}{} & \multicolumn{4}{l|}{} & \multicolumn{3}{l|}{} & \multicolumn{4}{l}{} \\
\multicolumn{1}{r|}{GPT-4o-2024-11-20} 
  & \cellcolor{red!50}2 & \underline{51.79} 
  & 64.91 & 55.81 & \textbf{44.44} 
  & 65.48 & \textbf{44.44} & 50.89 & \textbf{67.62} 
  & 29.76 & 48.78 & 70.24 & 34.48 & 58.02 & 14.81 & \underline{60.49} \\
\multicolumn{1}{r|}{Gemini-2.5-Pro} 
  & 5 & 49.08 
  & \underline{70.18} & 62.79 & \underline{37.04} 
  & \textbf{67.86} & 27.16 & 47.32 & 47.62 
  & 15.48 & 36.59 & \textbf{72.62} & \underline{41.38} & \underline{61.73} & \underline{34.57} & \textbf{65.43} \\
\multicolumn{1}{r|}{Claude-Sonnet-4-20250514} 
  & \cellcolor{red!10}3 & 50.73 
  & 59.65 & 55.81 & 33.33 
  & 61.90 & 33.33 & 52.68 & 56.19 
  & 35.71 & \textbf{57.32} & 71.43 & 20.69 & 60.49 & 30.86 & 51.85 \\
\multicolumn{1}{r|}{Qwen-Max-VL-latest} 
  & 4 & 50.53 
  & 52.63 & \underline{65.12} & 29.63 
  & 61.90 & \underline{41.98} & \underline{54.46} & 61.90 
  & \textbf{44.05} & 46.34 & 66.67 & 17.24 & 53.09 & \textbf{39.51} & 39.51 \\
\multicolumn{1}{r|}{Step-1o-turbo} 
  & \cellcolor{red!90}1 & \textbf{52.87} 
  & \textbf{75.00} & \textbf{70.83} & 21.05 
  & \underline{66.10} & 33.33 & \textbf{61.54} & 59.26 
  & 27.42 & \underline{55.93} & \underline{71.67} & \underline{41.38} & \textbf{67.27} & 18.52 & 56.60 \\
\multicolumn{1}{r|}{Doubao-seed-1-6-flash-250615} 
  & 2 & \underline{51.79} 
  & 59.65 & 48.84 & \underline{37.04} 
  & 54.76 & \textbf{44.44} & 53.57 & \underline{63.81} 
  & \underline{41.67} & 52.44 & 67.86 & \textbf{48.28} & 54.32 & \underline{34.57} & 48.10 \\
\hline
\rowcolor{cyan!8}
\multicolumn{1}{l|}{\textbf{\textit{Open-source Models}}} & \multicolumn{2}{l|}{} & \multicolumn{3}{l|}{} & \multicolumn{4}{l|}{} & \multicolumn{3}{l|}{} & \multicolumn{4}{l}{} \\
\multicolumn{1}{r|}{Phi-4-multimodal-instruct}                & 5  & 52.76    & 63.16    & 60.47    & 33.33    & 51.19    & 25.93    & 52.68    & \underline{65.71} & 26.19    & 40.24    & 70.24    & 24.14    & \textbf{66.67} & \underline{70.37} & 60.40    \\
\multicolumn{1}{r|}{Qwen2.5-VL-7B-Instruct}                   & 10 & 47.33    & \underline{66.67} & 60.47    & 25.93    & 63.10    & 25.93    & 50.89    & 51.43    & 47.56    & 47.56    & 66.67    & 13.79    & 34.57    & 25.93    & 43.21    \\
\multicolumn{1}{r|}{Qwen2.5-VL-72B-Instruct}                  & 4  & 54.90    & 59.65    & \underline{65.12} & 33.33    & 58.33    & \textbf{41.98} & \underline{63.39} & \textbf{67.62} & 48.78    & 48.78    & 73.81    & 17.24    & 59.26    & 37.04    & 48.15    \\
\multicolumn{1}{r|}{InternVL3-8B}                             & 6  & 52.18    & 56.14    & 60.47    & 25.93    & 59.52    & 30.86    & 58.04    & 56.19    & 56.10    & 51.22    & 71.43    & 20.69    & 54.32    & 53.09    & 50.62    \\
\multicolumn{1}{r|}{InternVL3-78B}                            & \cellcolor{red!10}3 & 55.38    & \underline{66.67} & \textbf{67.44} & \underline{44.44} & \underline{64.29} & 24.69    & \textbf{67.86} & 62.86    & \textbf{58.54} & \textbf{58.37} & \textbf{76.19} & 13.79    & 55.56    & 50.62    & 41.98    \\
\multicolumn{1}{r|}{Janus-Pro-7B}                             & 12 & 44.91    & 52.63    & 48.84    & 22.22    & 51.19    & 18.52    & 58.04    & 51.43    & 28.57    & 46.34    & 61.90    & 31.03    & \underline{60.49} & 33.33    & 37.00    \\
\multicolumn{1}{r|}{Chameleon-7B}                            & 15 & 38.22    & 36.84    & 37.21    & \underline{44.44} & 25.00    & 24.69    & 29.46    & 46.67    & 16.67    & 20.73    & 53.57    & 27.59    & 45.68    & \textbf{75.31} & 49.30    \\
\multicolumn{1}{r|}{PaliGemma-3B}                             & 17 & 24.25    & 19.30    & 37.21    & 22.22    & 30.95    & 35.80    & 18.75    & 13.33    & 11.90    & 21.95    & 47.62    & \textbf{65.52} & 17.28    & 16.05    & 16.05    \\
\multicolumn{1}{r|}{MiniCPM-V2.6}                             & 7  & 51.99    & 63.16    & 62.79    & 33.33    & \textbf{65.48} & \underline{40.74} & 49.11    & 49.52    & 46.43    & 48.78    & 66.67    & 41.38    & 58.02    & 46.91    & 45.68    \\
\multicolumn{1}{r|}{Ovis2-16B}                                & \cellcolor{red!90}1 & \textbf{59.17} & \textbf{68.42} & \textbf{67.44} & 29.63    & \underline{64.29} & 28.40    & 56.25    & \textbf{67.62} & \underline{58.33} & \underline{57.32} & 66.67    & \underline{51.72} & \underline{60.49} & 60.49    & \textbf{71.60} \\
\multicolumn{1}{r|}{Ovis-U1-3B}                              & 16 & 37.34    & 57.89    & 46.51    & 22.22    & 41.67    & 29.63    & 36.61    & 39.05    & 27.38    & 45.12    & 63.10    & 24.14    & 24.69    & 29.63    & 25.93    \\
\multicolumn{1}{r|}{Kimi-VL-A3B-Thinking}                     & \cellcolor{red!50}2 & \underline{56.84} & 59.65    & 60.47    & 25.93    & 61.90    & 38.27    & 58.04    & 63.81    & 45.24    & 50.00    & \textbf{76.19} & 48.28    & \textbf{66.67} & 51.85    & \underline{62.96} \\
\multicolumn{1}{r|}{Mimo-VL-7B-RL}                            & 9  & 48.59    & 61.40    & 58.14    & 29.63    & \underline{64.29} & 34.57    & 53.57    & 57.14    & 46.43    & 53.66    & \underline{75.00} & 10.34    & 50.62    & 17.28    & 33.33    \\
\multicolumn{1}{r|}{LLaVA-NeXT-7B-hf}                        & 14 & 38.31    & 28.07    & 46.51    & 18.52    & 35.71    & 25.93    & 39.29    & 52.38    & 29.76    & 37.80    & 59.52    & 27.59    & 55.56    & 25.93    & 29.63    \\
\multicolumn{1}{r|}{LLaVA-NeXT-13B-hf}                       & 13 & 39.28    & 31.58    & 44.19    & 33.33    & 42.86    & 30.86    & 40.18    & 46.67    & 40.48    & 41.46    & 50.00    & 34.48    & 44.44    & 34.57    & 24.69    \\
\multicolumn{1}{r|}{Skywork-R1V3}                             & 8  & 48.94    & 46.15    & 43.33    & \textbf{46.67} & 41.51    & 40.00    & 50.00    & 46.99    & 40.74    & 56.60    & 67.27    & 33.33    & 52.94    & 48.08    & 51.92    \\
\multicolumn{1}{r|}{mPLUG-OWL3}                              & 11 & 47.14    & 57.89    & 60.47    & 22.22    & 50.00    & 25.93    & 56.25    & 41.90    & 27.38    & 47.56    & 55.95    & 44.83    & 54.32    & 50.62    & 54.32    \\
\multicolumn{1}{r|}{XComposer-VL-7B}                         & 18 & 23.26    & 14.81    & 20.00    & 18.75    & 25.45    & 26.42    & 18.82    & 22.62    & 27.78    & 13.79    & 13.79    & 12.50    & 24.07    & 39.62    & 33.96    \\
\hline
\rowcolor{cyan!8}
\multicolumn{1}{l|}{\textbf{\textit{Fine-tuned Models}}} & \multicolumn{2}{l|}{} & \multicolumn{3}{l|}{} & \multicolumn{4}{l|}{} & \multicolumn{3}{l|}{} & \multicolumn{4}{l}{} \\
\multicolumn{1}{r|}{LLaVA-NeXT-13B} & \cellcolor{red!10}3  & 57.61        & 40.35        & \underline{60.47} & \underline{25.93} & 52.38        & \underline{45.68} & \underline{59.82} & 60.95        & 57.14        & 62.20        & 69.05        & \underline{37.93} & \underline{58.02} & 70.37        & 66.67        \\
\multicolumn{1}{r|}{Qwen-2.5-VL-7B} & \cellcolor{red!90}1          & \textbf{74.30} & \underline{63.16} & \textbf{65.12}  & \textbf{33.33}    & \textbf{69.05} & \textbf{75.31}    & \textbf{66.07}    & \textbf{72.38} & \textbf{76.19} & \textbf{82.93} & \textbf{83.33} & \textbf{55.17}    & \textbf{77.78} & \textbf{91.36} & \textbf{85.10} \\
\multicolumn{1}{r|}{Qwen-2.5-VL-3B} & \cellcolor{red!50}2 & \underline{66.44} & \textbf{73.68} & 55.81           & \textbf{33.33}    & \underline{59.52} & 34.57             & 57.14             & \underline{62.86} & \underline{66.67} & \underline{73.17} & \underline{82.14} & \textbf{55.17}    & \textbf{77.78} & \underline{90.12} & \underline{80.20}  \\
\hline
\rowcolor{cyan!8}
\multicolumn{1}{l|}{\textbf{\textit{Sim-to-Real Experiments}}} & \multicolumn{2}{l|}{} & \multicolumn{3}{l|}{} & \multicolumn{4}{l|}{} & \multicolumn{3}{l|}{} & \multicolumn{4}{l}{} \\
\multicolumn{1}{r|}{Qwen2.5-VL-7B}  
  & - & 47.77 & 50.00 & 55.56 & 11.11 & 83.33 & 50.00 & 27.78 & 55.56 & 61.11 & 44.44 & 82.35 & 10.53 & 38.89 & 64.71 & 27.70 \\
\multicolumn{1}{r|}{AirCop-7B} & -  & 67.41 & 50.00 & 77.78 & 11.11 & 83.33 & 88.89 & 77.78 & 77.78 & 77.78 & 94.44 & 82.35 & 31.58 & 50.00 & 76.47 & 50.00 \\
\hline
\end{tabular}
}
\end{table*}

\subsubsection{Quality Control.}
To ensure the reliability and effectiveness of our dataset for training and evaluating models on collaborative perception, we implement three key quality control measures: Standard Examination, Blind Filtering, and Human Refinement. 
\\ \indent
\textbf{Standard Examination.} 
This measure evaluates the quality of generated VQA pairs based on four criteria:
a) Required Content ensures all necessary information is included, flagging incomplete pairs for revision;
b) Format Consistency maintains uniform structure, wording, and presentation;
c) Answer Validity checks the correctness and relevance of options, filtering out incorrect ones and ensuring the correct option is present;
d) Question Length ensures questions are sufficiently detailed to avoid ambiguity.
\\ \indent
\textbf{Blind Filtering.}
This measure aims to remove questions answerable through common sense by using n MLLMs to predict answers without the multi-view image input. If all MLLMs answer correctly, the question is removed, filtering out invalid questions that can be answered solely through general perception knowledge \cite{zhao2025urbanvideo}.
\\ \indent
\textbf{Human Refinement.}
This step further ensures human modification of generated questions with issues such as:
a) Ambiguous Questions that are unclear, poorly framed, or irrelevant to the perception task;
b) Invalid Options that are missing correct answers, contain duplicates or mismatched content;
c) Incorrect Answers that are missing, contain multiple selections when only one is required, or are incorrectly chosen.
The entire human refinement process takes over 800 hours. Specific modified examples are provided in \textit{Appendix}.




\subsection{Benchmark Statistics} 
AirCopBench comprises 2,920 simultaneous multi-view images with various challenging perception degradation collected from real world and simulators, as shown in Fig.~\ref{statistics}b\&c. The dataset includes 14,610 questions, including both basic tasks like scene and object understanding, as well as advanced tasks like perception assessment and collaborative decision, as detailed in Fig.~\ref{statistics}a. 


\section{Experiments}
This section introduces the evaluated models and protocols, evaluates 40 mainstream MLLMs on AirCopBench, and summarizes results across model types. It also analyzes task correlations, sim-to-real potential, and error cases.

\subsection{Experimental Setup}
\subsubsection{Evaluation Metric.}
We integrate AirCopBench evaluation into the VLMEvalKit~\cite{duan2024vlmevalkit} framework, enabling capability assessments of new MLLMs. Specifically, we measure MLLM accuracy for each task type using question-answering accuracy.
\subsubsection{Baselines.}
We conduct zero-shot evaluations of 40 MLLMs on AirCopBench in Tab.~\ref{mllms_accuracy}. The baselines include both proprietary models, such as GPT-4o \cite{hurst2024gpt}, Gemini-2.5-Pro \cite{comanici2025gemini}, Claude-Sonnet-4~\cite{Claude3S}, and Qwen-Max-VL~\cite{Qwen-VL}, as well as open-source models capable of multi-image input, including LLaVA series \cite{liu2024llavanext}, Qwen-VL series \cite{bai2025qwen2, wang2024qwen2} and Phi series~\cite{abdin2024phi3, abdin2024phi4}, among others. Additional evaluated models include~\cite{chen2025janus, lu2023chameleon, yao2024minicpm, team2025kimi, Yue2025MiMoVLTR, ye2024mplugowl2, shen2025skywork, lu2024ovis,wang2025ovis, beyer2024paligemma, zhu2025internvl3}.


\subsection{Model Comparison}
From the quantitative results shown in Tab.~\ref{mllms_accuracy}, we have the following findings:

\begin{itemize}[left=0pt]
    \item \textbf{AirCopBench poses significant challenges to all MLLMs.}
    Both proprietary and open-source models perform poorly on aerial collaborative perception tasks, with the best model, Ovis-16B~\cite{lu2024ovis}, achieving only 59.17\% accuracy. This highlights the importance of AirCopBench in revealing the insufficient development of embodied perception and collaborative decision-making abilities in current MLLMs.

    \item \textbf{Leading open-source MLLMs match or exceed the best proprietary models on AirCopBench.}
    Models like Ovis2-16B~\cite{lu2024ovis} and Kimi-VL-A3B-Thinking~\cite{team2025kimi} match or slightly surpass top proprietary systems. However, the overall performance gap remains modest, and both open-source and proprietary MLLMs still struggle with multi-image understanding, underscoring the need for further progress in multi-image reasoning.
    
    


    \item \textbf{MLLMs demonstrate a clear bias across various question categories.}
    Current MLLMs perform well on tasks like Scene Description and Scene Comparison, but struggle with categories requiring domain knowledge and goal-oriented reasoning, such as Usability Assessment and When to Collaborate. This indicates a need for significant improvements in embodied perception assessment and adaptive collaboration decision-making.
\end{itemize}

\subsection{Correlation Analysis}
\begin{figure}[t]
    \centering
    \includegraphics[width=1\linewidth]
    {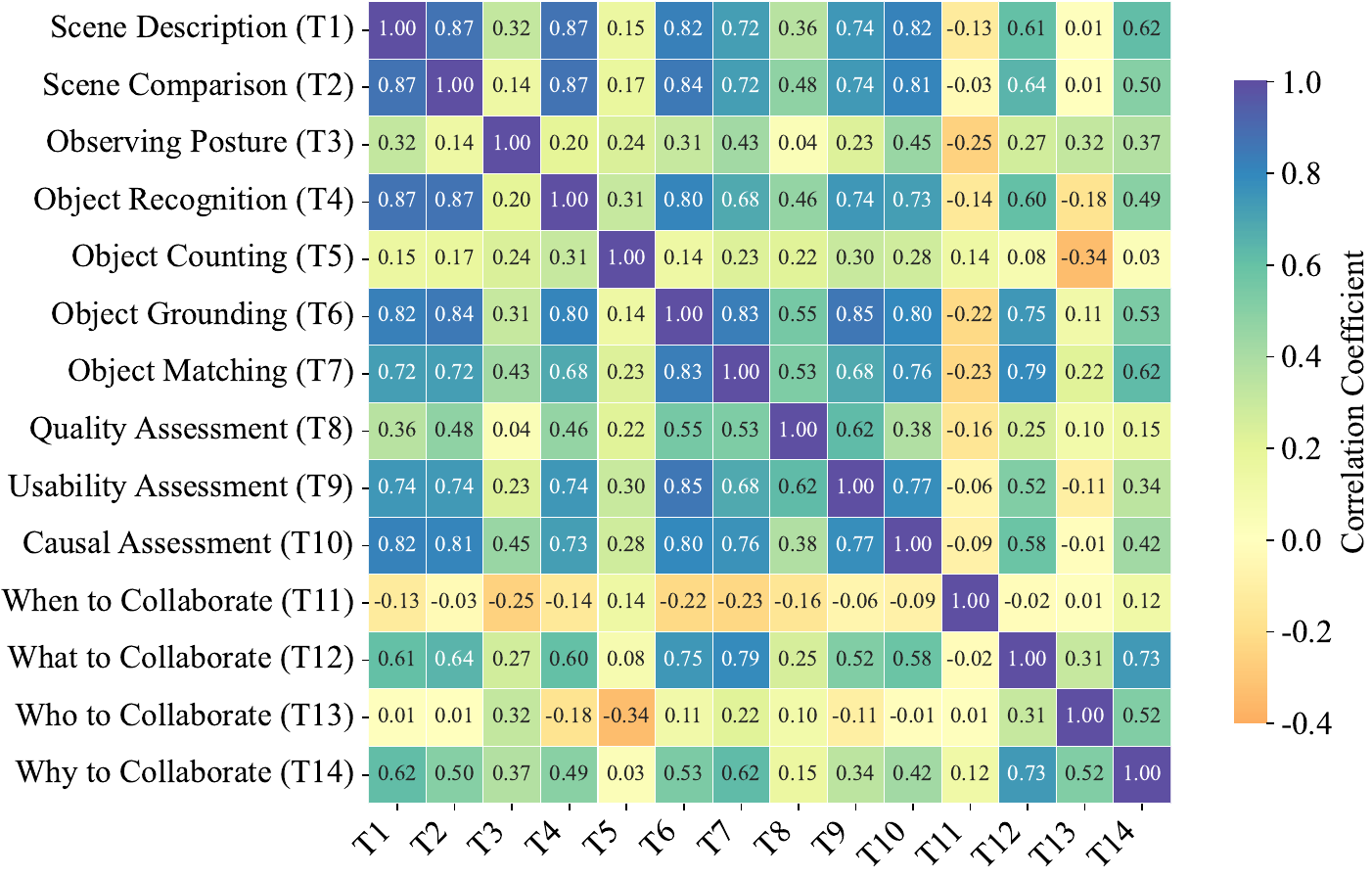}
    \caption{Correlation coefficients of MLLMs' performance across all tasks, with higher values indicating greater similarity in the cognitive abilities required by the two tasks.}
    \label{correlation}
    \vspace{-0.3cm}
\end{figure}
To explore the relationships between tasks and their required cognitive abilities, we compute pairwise correlations of MLLMs' accuracy (Tab.~\ref{comparison}) on each task. Similar performance across models on two tasks suggests shared cognitive abilities. From Fig.~\ref{correlation}, we draw the following conclusions:
\begin{itemize}[left=0pt]
    \item \textbf{Causal Assessment exhibits a high correlation with almost all other tasks.} It suggests that understanding and inferring the causality of perception degradation is crucial for various cognitive processes. This finding implies that causal assessment could be a key factor in the development of embodied cognition in collaborative perception.
    \item \textbf{Object Matching has a high correlation with both Object Recognition and Object Grounding tasks.} This observation aligns with prior knowledge that effective multi-view image understanding relies on strong single-image perceptual capabilities.
    \item \textbf{Quality Assessment exhibits moderate correlations with multiple tasks.} This suggests that quality evaluation requires integrating Scene Understanding, Object Recognition, and Object Matching abilities, making it a highly composite decision‐making task.
\end{itemize}

\subsection{Supervised Fine-Tuning}
We fine-tune two representative MLLMs, Qwen-2.5-VL (7B and 3B)~\cite{bai2025qwen2} and LLaVA-NeXT-13B~\cite{liu2024llavanext}, using our curated instruction dataset within the LLaMA-Factory framework~\cite{zheng2024llamafactory}, with default hyperparameters for three epochs. As shown in Tab.~\ref{mllms_accuracy}, Qwen-2.5-VL-7B achieves 74.30\% accuracy (+26.97), Qwen-2.5-VL-3B reaches 66.44\% (+19.11), and LLaVA-NeXT-13B improves to 57.61\% (+19.30), demonstrating that domain-specific SFT substantially enhances MLLM performance in collaborative perception.



\subsection{Sim-to-Real}
We evaluate the generalization of models trained on simulated data to real-world UAV imagery. As shown in Tab.~\ref{mllms_accuracy}, we compare the open-source Qwen2.5-VL-7B~\cite{bai2025qwen2} and our AirCop-7B model fine-tuned on simulator data. Qwen2.5-VL-7B achieves 47.77\% overall accuracy, performing well in Object Recognition (83.33\%) but poorly in Posture (11.11\%) and Collaboration (40\%). After fine-tuning, AirCop-7B reaches 67.41\% (+19.64), with notable gains in Object Grounding (77.78\%, +50.00), Object Counting (88.89\%, +38.89), and Usability Assessment (94.44\%, +53.20). These results confirm that simulator-based fine-tuning significantly enhances sim-to-real transfer, improving generalization to real UAV imagery and validating the effectiveness of our simulated dataset.

\subsection{Error Analysis}
The errors in MLLM reasoning primarily stem from three causes, as shown in Fig.~\ref{error}:
\begin{figure}
    \centering
    \includegraphics[width=1\linewidth]{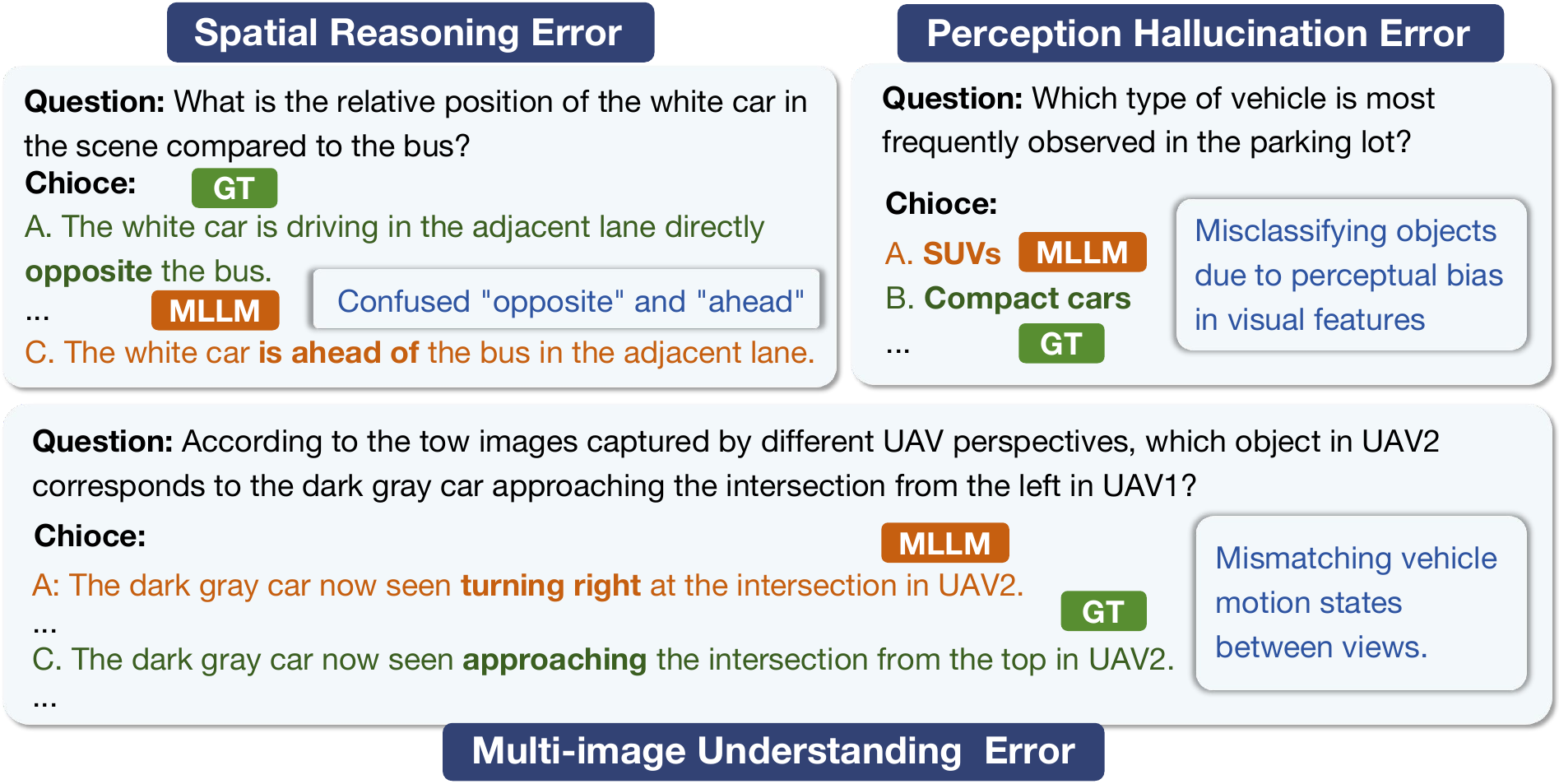}
    \caption{Examples of three common errors in MLLM reasoning during aerial collaborative perception tasks.}
    \label{error}
    \vspace{-0.3cm}
\end{figure}
\begin{itemize}[left=0pt]
    \item \textbf{Perception hallucination errors.} 
    MLLMs misidentify or fail to recognize objects due to perception hallucination, where visual input is incorrectly processed, leading to false or exaggerated perceptions of objects.
    
    \item \textbf{Spatial reasoning errors.}
    MLLMs struggle with spatial reasoning, causing errors in interpreting object relationships, such as incorrect positioning or orientation of objects in space \cite{zhang2025point, zha2025enable}.
    
    \item \textbf{Multi-image understanding errors.} 
    MLLMs face challenges in multi-image understanding, resulting in errors when comparing, matching, or integrating information across images, leading to incorrect logical inferences.
\end{itemize}
Perceptual and spatial errors hinder Object Recognition and Grounding, while multi-image reasoning limits Scene Comparison, Object Matching, and Collaborative Decision; other tasks show varied failure modes (see \textit{Appendix}).


\section{Conclusion}

In this work, we propose AirCopBench, a benchmark for multi-UAV collaborative perception with challenging degradation, featuring 2.9k+ multi-view images and 14.6k+ questions. We evaluate 40 popular MLLMs on scene understanding, object understanding, perception assessment, and collaborative decision. The best-performing models achieve only 59.17\% a ccuracy, highlighting challenges in multi-view collaborative perception. Their performance also shows biases across task types. Fine-tuned MLLMs with simulated data improves their performance on real-world tasks. Future work will extend to more diverse real-world data and explore efficient MLLM architectures for practical multi-UAV deployment.

\section{Acknowledgments}
This paper was supported by the Natural Science Foundation of China under Grant 62371269, Shenzhen Low-Altitude Airspace Strategic Program Portfolio Z253061 and Meituan Academy of Robotics Shenzhen. Sponsored by Tsinghua University-Toyota Research Center.

\bibliography{aaai2026}

@article{yeh2025seeing,
  title={Seeing from another perspective: Evaluating multi-view understanding in mllms},
  author={Yeh, Chun-Hsiao and Wang, Chenyu and Tong, Shengbang and Cheng, Ta-Ying and Wang, Ruoyu and Chu, Tianzhe and Zhai, Yuexiang and Chen, Yubei and Gao, Shenghua and Ma, Yi},
  journal={arXiv preprint arXiv:2504.15280},
  year={2025}
}

@inproceedings{zhou2025urbench,
  title={Urbench: A comprehensive benchmark for evaluating large multimodal models in multi-view urban scenarios},
  author={Zhou, Baichuan and Yang, Haote and Chen, Dairong and Ye, Junyan and Bai, Tianyi and Yu, Jinhua and Zhang, Songyang and Lin, Dahua and He, Conghui and Li, Weijia},
  booktitle={Proceedings of the AAAI Conference on Artificial Intelligence},
  volume={39},
  number={10},
  pages={10707--10715},
  year={2025}
}

@article{liu2023robust,
  title={Robust multi-drone multi-target tracking to resolve target occlusion: A benchmark},
  author={Liu, Zhihao and Shang, Yuanyuan and Li, Timing and Chen, Guanlin and Wang, Yu and Hu, Qinghua and Zhu, Pengfei},
  journal={IEEE Transactions on Multimedia},
  volume={25},
  pages={1462--1476},
  year={2023},
  publisher={IEEE}
}

@article{wang2024muirbench,
  title={Muirbench: A comprehensive benchmark for robust multi-image understanding},
  author={Wang, Fei and Fu, Xingyu and Huang, James Y and Li, Zekun and Liu, Qin and Liu, Xiaogeng and Ma, Mingyu Derek and Xu, Nan and Zhou, Wenxuan and Zhang, Kai and others},
  journal={arXiv preprint arXiv:2406.09411},
  year={2024}
}

@article{wang2024drones,
  title={Drones help drones: A collaborative framework for multi-drone object trajectory prediction and beyond},
  author={Wang, Zhechao and Cheng, Peirui and Chen, Minxing and Tian, Pengju and Wang, Zhirui and Li, Xinming and Yang, Xue and Sun, Xian},
  journal={Advances in Neural Information Processing Systems},
  volume={37},
  pages={64604--64628},
  year={2024}
}

@article{sunderraman2024uav3d,
  title={UAV3D: A Large-scale 3D Perception Benchmark for Unmanned Aerial Vehicles},
  author={Sunderraman, Rajshekhar and Ji, Jonathan Shihao and others},
  journal={Advances in Neural Information Processing Systems},
  volume={37},
  pages={55425--55442},
  year={2024}
}

@article{tian2024ucdnet,
  title={Ucdnet: Multi-uav collaborative 3d object detection network by reliable feature mapping},
  author={Tian, Pengju and Wang, Zhirui and Cheng, Peirui and Wang, Yuchao and Wang, Zhechao and Zhao, Liangjin and Yan, Menglong and Yang, Xue and Sun, Xian},
  journal={IEEE Transactions on Geoscience and Remote Sensing},
  year={2024},
  publisher={IEEE}
}

@inproceedings{dosovitskiy2017carla,
  title={CARLA: An open urban driving simulator},
  author={Dosovitskiy, Alexey and Ros, German and Codevilla, Felipe and Lopez, Antonio and Koltun, Vladlen},
  booktitle={Conference on robot learning},
  pages={1--16},
  year={2017},
  organization={PMLR}
}

@inproceedings{shah2017airsim,
  title={Airsim: High-fidelity visual and physical simulation for autonomous vehicles},
  author={Shah, Shital and Dey, Debadeepta and Lovett, Chris and Kapoor, Ashish},
  booktitle={Field and service robotics: Results of the 11th international conference},
  pages={621--635},
  year={2017},
  organization={Springer}
}

@article{zhao2025urbanvideo,
  title={Urbanvideo-bench: Benchmarking vision-language models on embodied intelligence with video data in urban spaces},
  author={Zhao, Baining and Fang, Jianjie and Dai, Zichao and Wang, Ziyou and Zha, Jirong and Zhang, Weichen and Gao, Chen and Wang, Yue and Cui, Jinqiang and Chen, Xinlei and others},
  journal={arXiv preprint arXiv:2503.06157},
  year={2025}
}

@inproceedings{saini2025advancing,
  title={Advancing open-set object detection in remote sensing using multimodal large language model},
  author={Saini, Nandini and Dubey, Ashudeep and Das, Debasis and Chattopadhyay, Chiranjoy},
  booktitle={Proceedings of the Winter Conference on Applications of Computer Vision},
  pages={451--458},
  year={2025}
}

@article{kil2024mllm,
  title={Mllm-compbench: A comparative reasoning benchmark for multimodal llms},
  author={Kil, Jihyung and Mai, Zheda and Lee, Justin and Chowdhury, Arpita and Wang, Zihe and Cheng, Kerrie and Wang, Lemeng and Liu, Ye and Chao, Wei-Lun Harry},
  journal={Advances in Neural Information Processing Systems},
  volume={37},
  pages={28798--28827},
  year={2024}
}

@article{hurst2024gpt,
  title={Gpt-4o system card},
  author={Hurst, Aaron and Lerer, Adam and Goucher, Adam P and Perelman, Adam and Ramesh, Aditya and Clark, Aidan and Ostrow, AJ and Welihinda, Akila and Hayes, Alan and Radford, Alec and others},
  journal={arXiv preprint arXiv:2410.21276},
  year={2024}
}

@article{wang2024qwen2,
  title={Qwen2-vl: Enhancing vision-language model's perception of the world at any resolution},
  author={Wang, Peng and Bai, Shuai and Tan, Sinan and Wang, Shijie and Fan, Zhihao and Bai, Jinze and Chen, Keqin and Liu, Xuejing and Wang, Jialin and Ge, Wenbin and others},
  journal={arXiv preprint arXiv:2409.12191},
  year={2024}
}

@article{bai2025qwen2,
  title={Qwen2. 5-vl technical report},
  author={Bai, Shuai and Chen, Keqin and Liu, Xuejing and Wang, Jialin and Ge, Wenbin and Song, Sibo and Dang, Kai and Wang, Peng and Wang, Shijie and Tang, Jun and others},
  journal={arXiv preprint arXiv:2502.13923},
  year={2025}
}

@article{lu2023chameleon,
  title={Chameleon: Plug-and-play compositional reasoning with large language models},
  author={Lu, Pan and Peng, Baolin and Cheng, Hao and Galley, Michel and Chang, Kai-Wei and Wu, Ying Nian and Zhu, Song-Chun and Gao, Jianfeng},
  journal={Advances in Neural Information Processing Systems},
  volume={36},
  pages={43447--43478},
  year={2023}
}

@article{beyer2024paligemma,
  title={Paligemma: A versatile 3b vlm for transfer},
  author={Beyer, Lucas and Steiner, Andreas and Pinto, Andr{\'e} Susano and Kolesnikov, Alexander and Wang, Xiao and Salz, Daniel and Neumann, Maxim and Alabdulmohsin, Ibrahim and Tschannen, Michael and Bugliarello, Emanuele and others},
  journal={arXiv preprint arXiv:2407.07726},
  year={2024}
}

@article{yao2024minicpm,
  title={Minicpm-v: A gpt-4v level mllm on your phone},
  author={Yao, Yuan and Yu, Tianyu and Zhang, Ao and Wang, Chongyi and Cui, Junbo and Zhu, Hongji and Cai, Tianchi and Li, Haoyu and Zhao, Weilin and He, Zhihui and others},
  journal={arXiv preprint arXiv:2408.01800},
  year={2024}
}

@misc{liu2024llavanext,
    title={LLaVA-NeXT: Improved reasoning, OCR, and world knowledge},
    url={https://llava-vl.github.io/blog/2024-01-30-llava-next/},
    author={Liu, Haotian and Li, Chunyuan and Li, Yuheng and Li, Bo and Zhang, Yuanhan and Shen, Sheng and Lee, Yong Jae},
    month={January},
    year={2024}
}

@article{lu2024ovis,
  title={Ovis: Structural embedding alignment for multimodal large language model},
  author={Lu, Shiyin and Li, Yang and Chen, Qing-Guo and Xu, Zhao and Luo, Weihua and Zhang, Kaifu and Ye, Han-Jia},
  journal={arXiv preprint arXiv:2405.20797},
  year={2024}
}

@inproceedings{ye2024mplugowl2,
  title={mplug-owl2: Revolutionizing multi-modal large language model with modality collaboration},
  author={Ye, Qinghao and Xu, Haiyang and Ye, Jiabo and Yan, Ming and Hu, Anwen and Liu, Haowei and Qian, Qi and Zhang, Ji and Huang, Fei},
  booktitle={Proceedings of the ieee/cvf conference on computer vision and pattern recognition},
  pages={13040--13051},
  year={2024}
}

@article{abdin2024phi3,
  title={Phi-3 technical report: A highly capable language model locally on your phone},
  author={Abdin, Marah and Aneja, Jyoti and Awadalla, Hany and Awadallah, Ahmed and Awan, Ammar Ahmad and Bach, Nguyen and Bahree, Amit and Bakhtiari, Arash and Bao, Jianmin and Behl, Harkirat and others},
  journal={arXiv preprint arXiv:2404.14219},
  year={2024}
}

@article{abdin2024phi4,
  title={Phi-4 technical report},
  author={Abdin, Marah and Aneja, Jyoti and Behl, Harkirat and Bubeck, S{\'e}bastien and Eldan, Ronen and Gunasekar, Suriya and Harrison, Michael and Hewett, Russell J and Javaheripi, Mojan and Kauffmann, Piero and others},
  journal={arXiv preprint arXiv:2412.08905},
  year={2024}
}

@inproceedings{duan2024vlmevalkit,
  title={Vlmevalkit: An open-source toolkit for evaluating large multi-modality models},
  author={Duan, Haodong and Yang, Junming and Qiao, Yuxuan and Fang, Xinyu and Chen, Lin and Liu, Yuan and Dong, Xiaoyi and Zang, Yuhang and Zhang, Pan and Wang, Jiaqi and others},
  booktitle={Proceedings of the 32nd ACM International Conference on Multimedia},
  pages={11198--11201},
  year={2024}
}

@article{comanici2025gemini,
  title={Gemini 2.5: Pushing the frontier with advanced reasoning, multimodality, long context, and next generation agentic capabilities},
  author={Comanici, Gheorghe and Bieber, Eric and Schaekermann, Mike and Pasupat, Ice and Sachdeva, Noveen and Dhillon, Inderjit and Blistein, Marcel and Ram, Ori and Zhang, Dan and Rosen, Evan and others},
  journal={arXiv preprint arXiv:2507.06261},
  year={2025}
}

@article{Qwen-VL,
  title={Qwen-VL: A Versatile Vision-Language Model for Understanding, Localization, Text Reading, and Beyond},
  author={Bai, Jinze and Bai, Shuai and Yang, Shusheng and Wang, Shijie and Tan, Sinan and Wang, Peng and Lin, Junyang and Zhou, Chang and Zhou, Jingren},
  journal={arXiv preprint arXiv:2308.12966},
  year={2023}
}

@article{zhu2025internvl3,
  title={Internvl3: Exploring advanced training and test-time recipes for open-source multimodal models},
  author={Zhu, Jinguo and Wang, Weiyun and Chen, Zhe and Liu, Zhaoyang and Ye, Shenglong and Gu, Lixin and Tian, Hao and Duan, Yuchen and Su, Weijie and Shao, Jie and others},
  journal={arXiv preprint arXiv:2504.10479},
  year={2025}
}

@article{chen2025janus,
  title={Janus-pro: Unified multimodal understanding and generation with data and model scaling},
  author={Chen, Xiaokang and Wu, Zhiyu and Liu, Xingchao and Pan, Zizheng and Liu, Wen and Xie, Zhenda and Yu, Xingkai and Ruan, Chong},
  journal={arXiv preprint arXiv:2501.17811},
  year={2025}
}

@article{wang2025ovis,
  title={Ovis-U1 Technical Report},
  author={Wang, Guo-Hua and Zhao, Shanshan and Zhang, Xinjie and Cao, Liangfu and Zhan, Pengxin and Duan, Lunhao and Lu, Shiyin and Fu, Minghao and Chen, Xiaohao and Zhao, Jianshan and others},
  journal={arXiv preprint arXiv:2506.23044},
  year={2025}
}

@article{team2025kimi,
  title={Kimi-vl technical report},
  author={Team, Kimi and Du, Angang and Yin, Bohong and Xing, Bowei and Qu, Bowen and Wang, Bowen and Chen, Cheng and Zhang, Chenlin and Du, Chenzhuang and Wei, Chu and others},
  journal={arXiv preprint arXiv:2504.07491},
  year={2025}
}

@article{Yue2025MiMoVLTR,
  title={MiMo-VL Technical Report},
  author={Xiaomi LLM-Core Team Zihao Yue and Zhenrui Lin and Yi-Hao Song and Weikun Wang and Shu-Qin Ren and Shuhao Gu and Shi-Guang Li and Peidian Li and Liang Zhao and Lei Li and Kainan Bao and Hao Tian and Hailin Zhang and Gang Wang and Dawei Zhu and Cici and Chenhong He and Bowen Ye and Bowen Shen and Zihan Zhang and Zi-Ang Jiang and Zhixian Zheng and Zhichao Song and others},
  journal={ArXiv},
  year={2025},
  volume={abs/2506.03569},
  url={https://api.semanticscholar.org/CorpusID:279155294}
}

@inproceedings{liu2020when2com,
  title={When2com: Multi-agent perception via communication graph grouping},
  author={Liu, Yen-Cheng and Tian, Junjiao and Glaser, Nathaniel and Kira, Zsolt},
  booktitle={Proceedings of the IEEE/CVF Conference on computer vision and pattern recognition},
  pages={4106--4115},
  year={2020}
}

@inproceedings{hu2022where2comm,
  author    = {Hu, Yue and Fang, Shaoheng and Lei, Zixing and Zhong, Yiqi and Chen, Siheng},
  title     = {Where2comm: Communication-Efficient Collaborative Perception via Spatial Confidence Maps},
  booktitle = {Advances in Neural Information Processing Systems},
  month     = {November},  
  year      = {2022}
}

@article{zhang2024mm,
  title={Mm-llms: Recent advances in multimodal large language models},
  author={Zhang, Duzhen and Yu, Yahan and Dong, Jiahua and Li, Chenxing and Su, Dan and Chu, Chenhui and Yu, Dong},
  journal={arXiv preprint arXiv:2401.13601},
  year={2024}
}

@article{gao2024embodiedcity,
  title={Embodiedcity: A benchmark platform for embodied agent in real-world city environment},
  author={Gao, Chen and Zhao, Baining and Zhang, Weichen and Mao, Jinzhu and Zhang, Jun and Zheng, Zhiheng and Man, Fanhang and Fang, Jianjie and Zhou, Zile and Cui, Jinqiang and others},
  journal={arXiv preprint arXiv:2410.09604},
  year={2024}
}

@article{shen2025skywork,
  title={Skywork-R1V3 Technical Report},
  author={Shen, Wei and Pei, Jiangbo and Peng, Yi and Song, Xuchen and Liu, Yang and Peng, Jian and Sun, Haofeng and Hao, Yunzhuo and Wang, Peiyu and Zhou, Yahui},
  journal={arXiv preprint arXiv:2507.06167},
  year={2025}
}

@inproceedings{zheng2024llamafactory,
  title={LlamaFactory: Unified Efficient Fine-Tuning of 100+ Language Models},
  author={Yaowei Zheng and Richong Zhang and Junhao Zhang and Yanhan Ye and Zheyan Luo and Zhangchi Feng and Yongqiang Ma},
  booktitle={Proceedings of the 62nd Annual Meeting of the Association for Computational Linguistics (Volume 3: System Demonstrations)},
  address={Bangkok, Thailand},
  publisher={Association for Computational Linguistics},
  year={2024},
  url={http://arxiv.org/abs/2403.13372}
}

@techreport{Claude3S,
  author = {Anthropic},
  title = {Claude 3.7 Sonnet “Extended Thinking” System Card},
  year = {2025},
  url ={https://api.semanticscholar.org/CorpusID:276612236},
}

@article{zang2025contextual,
  title={Contextual object detection with multimodal large language models},
  author={Zang, Yuhang and Li, Wei and Han, Jun and Zhou, Kaiyang and Loy, Chen Change},
  journal={International Journal of Computer Vision},
  volume={133},
  number={2},
  pages={825--843},
  year={2025},
  publisher={Springer}
}

@article{zha2025enable,
  title={How to enable llm with 3d capacity? a survey of spatial reasoning in llm},
  author={Zha, Jirong and Fan, Yuxuan and Yang, Xiao and Gao, Chen and Chen, Xinlei},
  journal={arXiv preprint arXiv:2504.05786},
  year={2025}
}

@article{bai2025hallucination,
  title={Hallucination at a Glance: Controlled Visual Edits and Fine-Grained Multimodal Learning},
  author={Bai, Tianyi and Fan, Yuxuan and Qiu, Jiantao and Sun, Fupeng and Song, Jiayi and Han, Junlin and Liu, Zichen and He, Conghui and Zhang, Wentao and Yuan, Binhang},
  journal={arXiv preprint arXiv:2506.07227},
  year={2025}
}

@article{huang2025mllm,
  title={MLLM-For3D: Adapting Multimodal Large Language Model for 3D Reasoning Segmentation},
  author={Huang, Jiaxin and Chen, Runnan and Li, Ziwen and Gao, Zhengqing and He, Xiao and Guo, Yandong and Gong, Mingming and Liu, Tongliang},
  journal={arXiv preprint arXiv:2503.18135},
  year={2025}
}

@article{cheng2025evaluating,
  title={Evaluating mllms with multimodal multi-image reasoning benchmark},
  author={Cheng, Ziming and Xu, Binrui and Gong, Lisheng and Song, Zuhe and Zhou, Tianshuo and Zhong, Shiqi and Ren, Siyu and Chen, Mingxiang and Meng, Xiangchao and Zhang, Yuxin and others},
  journal={arXiv preprint arXiv:2506.04280},
  year={2025}
}

@article{zhang2025point,
  title={The Point, the Vision and the Text: Does Point Cloud Boost Spatial Reasoning of Large Language Models?},
  author={Zhang, Weichen and Peng, Ruiying and Gao, Chen and Fang, Jianjie and Zeng, Xin and Li, Kaiyuan and Wang, Ziyou and Cui, Jinqiang and Wang, Xin and Chen, Xinlei and others},
  journal={arXiv preprint arXiv:2504.04540},
  year={2025}
}

@article{zha2024diffusion,
  title={Diffusion-based filter for fast and accurate collaborative tracking with low data transmission},
  author={Zha, Jirong and Zhou, Nan and Liu, Zhenyu and Sun, Tao and Chen, Xinlei},
  journal={Authorea Preprints},
  year={2024},
  publisher={Authorea}
}

@article{caffagni2024revolution,
  title={The revolution of multimodal large language models: a survey},
  author={Caffagni, Davide and Cocchi, Federico and Barsellotti, Luca and Moratelli, Nicholas and Sarto, Sara and Baraldi, Lorenzo and Cornia, Marcella and Cucchiara, Rita},
  journal={arXiv preprint arXiv:2402.12451},
  year={2024}
}

@inproceedings{fu2024blink,
  title={Blink: Multimodal large language models can see but not perceive},
  author={Fu, Xingyu and Hu, Yushi and Li, Bangzheng and Feng, Yu and Wang, Haoyu and Lin, Xudong and Roth, Dan and Smith, Noah A and Ma, Wei-Chiu and Krishna, Ranjay},
  booktitle={European Conference on Computer Vision},
  pages={148--166},
  year={2024},
  organization={Springer}
}

@inproceedings{lin2025mcop,
  title={MCOP: Multi-UAV Collaborative Occupancy Prediction},
  author={Lin, Zefu and Chen, Wenbo and Jin, Xiaojuan and Yang, Yuran and Fan, Lue and Zhang, Yixin and Zhang, Yufeng and Zhang, Zhaoxiang},
  booktitle={Proceedings of the IEEE/CVF International Conference on Computer Vision},
  pages={27242--27251},
  year={2025}
}

@inproceedings{xiao2025uav,
  title={Uav-on: A benchmark for open-world object goal navigation with aerial agents},
  author={Xiao, Jianqiang and Sun, Yuexuan and Shao, Yixin and Gan, Boxi and Liu, Rongqiang and Wu, Yanjin and Guan, Weili and Deng, Xiang},
  booktitle={Proceedings of the 33rd ACM International Conference on Multimedia},
  pages={13023--13029},
  year={2025}
}

@article{guo2025bedi,
  title={Bedi: A comprehensive benchmark for evaluating embodied agents on uavs},
  author={Guo, Mingning and Wu, Mengwei and He, Jiarun and Li, Shaoxian and Li, Haifeng and Tao, Chao},
  journal={arXiv preprint arXiv:2505.18229},
  year={2025}
}

@article{chen2025large,
  title={When Large Language Models Meet UAVs: How Far Are We?},
  author={Chen, Yihua and Que, Xingle and Zhang, Jiashuo and Chen, Ting and Li, Guangshun and Chen, Jiachi},
  journal={arXiv preprint arXiv:2509.12795},
  year={2025}
}

\end{document}